\documentclass[conference]{IEEEtran}

\title{Faster Q-Learning Algorithms for Restless Bandits }

\author{Parvish  Kakarapalli \ \ \ & \ \ \ Devendra Kayande \ \ \ & \ \ \  Rahul Meshram   \\
 \ \ IT Deptt. \ \ \ \ \ \ & \ \ \ \ \ \ \ \   ECE Deptt. \ \ \ \ \ \ &  \ \ \ \   EE  Deptt.  \\
IIIT Allahabad \ \ \ \ & \ \ \ \ \ \ \  IIIT Allahabad  \ \ \ \ \  \ & \ \ \  \  IIT Madras
}

\usepackage{amsmath,amscd,amssymb,amsgen,amsfonts,amsbsy}
\usepackage{mathrsfs}
\usepackage[utf8x]{inputenc}
\usepackage{color}
\usepackage{textcomp}
\usepackage{float}
\usepackage{latexsym,graphicx}
\usepackage{mathtools}
\usepackage{breqn}
\usepackage{tikz}
\usepackage{breqn}
\usetikzlibrary{automata,arrows,positioning,calc}
\usepackage{ragged2e}
\usepackage{url}
\usepackage{cite}
\usepackage{algorithmic}
\usepackage[ruled,lined]{algorithm2e}
 \usepackage{blkarray}

\newcommand{\remove}[1]{}

\newcommand{\comments}[1]{}

\def\BibTeX{{\rm B\kern-.05em{\sc i\kern-.025em b}\kern-.08em
    T\kern-.1667em\lower.7ex\hbox{E}\kern-.125emX}}
\begin{document} 

\maketitle

\begin{abstract}
    We study the Whittle index learning algorithm for restless multi-armed bandits (RMAB).  We first present  Q-learning algorithm and its variants---speedy Q-learning (SQL), generalized speedy Q-learning (GSQL) and phase Q-learning (PhaseQL). We also discuss exploration policies---$\epsilon$-greedy and Upper confidence bound (UCB). We extend the study of Q-learning and its variants with UCB policy. We illustrate using numerical example that Q-learning with UCB exploration policy has faster convergence and PhaseQL with UCB have fastest convergence rate. We next  extend the study of Q-learning variants for index learning to RMAB. The algorithm of index  learning is two-timescale variant of stochastic approximation, on slower timescale we update index learning scheme and on faster timescale we update Q-learning assuming fixed index value.  We study constant stepsizes two timescale stochastic approximation algorithm. 


    We describe the performance of our algorithms using numerical example. It illustrate that index learning with Q learning with UCB has faster convergence that $\epsilon$ greedy. Further, PhaseQL (with UCB and $\epsilon$ greedy) has  the best convergence than other Q-learning algorithms.

\end{abstract}

\section{Introduction}

Restless multi-armed bandit (RMAB) is a class of sequential decision problem and it has applications to resource allocation problems---opportunistic communication systems \cite{Borkar17b}, queueing  systems \cite{Glazebrook05}, machine repair and maintenance \cite{Glazebrook2006}, healthcare \cite{Mate21}, recommendation systems \cite{Meshram2016,Meshram17}, and multi-target tracking systems  \cite{Ny2008,Whittle88}. Other applications of RMAB are given in \cite{NinoMora23}.


RMAB consists of $N$ independent arms, i.e, independent Markov processes and state of arms is evolving at each time step. The agent has to determine at each time step which sequence to act or play an arm so that total expected cumulative reward is maximized. In \cite{Whittle88}, author proposed the index based policy and it is also referred to as Whittle index policy.  Popularity of Whittle index policy is due to that it is shown to be near optimal,  \cite{Weber90}. 
RMAB is also called as weakly couple Markov decision processes. The Whittle index computation involves decoupling these independent Markov decision processes.
In the  index policy, an arm with the highest index based on the state of each arm is played at each time step. To use the  index based policy, the agent need to know the model, i.e., transition probabilities from a state to another state and reward matrix. In this paper we assume that model is unknown. The agent employs learning algorithms. 

Q-learning algorithm is extensively studied for Markov decision processes (MDPs) when the model is unknown to the agent, \cite{Watkins1989,Borkar2000,Borkar08}. The convergence analysis of Q-learning algorithm is provided using stochastic approximation scheme, \cite{Satinder2000,Borkar2000}.  Further, the rate of convergence depends on structure of model and an exploration policy that is employed for action selection during Q-learning updates, for example $\epsilon-$greedy policy.  
Recently, Q-learning algorithm is studied for RMAB  to learn the index of each arm for given state, \cite{Fu2019,Avrachenkov2022}. The index is learned via two-timescales stochastic approximations scheme and decreasing learning rate is assumed for the stronger convergence results.  In Q-learning based index learning algorithm, there are no  structural assumptions that are made on model except---1) assumptions on learning rate (stepsizes), 2) each state-action pair visited infinitely often. With Q-learning approach, one directly learns the index for each arm separately, and this can be learnt independently  using parallel  processing.  Once the index is obtained from learning scheme for all arms and all states, the Whittle index policy can be utilized to compute the expected cumulative reward. 



In this paper, we study faster variants of Q-learning algorithm for RMAB---speedy Q learning (SQL), generalized speedy Q-learning (GSQL), phase Q-learning (PhaseQL).
In classical Q-learning algorithm, $\epsilon$-greedy policy is used as action selection scheme as exploration-exploitation policy. We introduced upper confidence bound (UCB) scheme as exploration-exploitation policy along with different Q-learning variants.

\subsection{Related Work}



Recently, Q-learning algorithm is  studied for restless bandits, \cite{Fu2019,Robledo2021,Avrachenkov2022, Xiong2023}. Neural-network based model is proposed to learn Whittle index for restless bandits in \cite{Nakhleh2021}.  The neural networks approach is combined with Q-learning in \cite{Robledo2022} and they studied QWI and QWINN. 
Web crawling problem is modeled as RMAB and deep RL based learning is studied in  \cite{Avrachenkov2021b}. The model of deep RL is based on batch gradient descent, and in the work of \cite{Pagare2023}, it is studied for deep RL with full gradient for average criteria, and discounted criteria in \cite{Avrachenkov2021}. Q-learning is model free learning algorithm which directly learn Q-estimeate and index.


UCB based leaning algorithms for rested and restless bandits are studied in \cite{Tekin2012}. Upper confidence bound reinforcement learning (UCRL) algorithm is analyzed for RMAB in \cite{Ortner2014}, where model is estimated and regret bounds are provided. The UCB variant of learning algorithm for RMAB is studied in \cite{Wang2020}.
Thompson sampling algorithm for hidden Markov restless bandits is proposed in \cite{Meshram2016,Meshram17} and the analysis of regret bound is provided by \cite{Jung2019},\cite{Wang2023}. 
UCB and Thompson sampling algorithms studied for RMAB are model based learning approaches. Q-learning with UCB is studied in \cite{Jin2018}, where they improved efficiency and regret bounds.
a
Variants of Q-learning proposed to speed up convergence rate. Here we mention few variants--- 1) speedy Q learning (SQL) is introduced in \cite{Azar2011}, 2) Generalized speedy Q learning (GSQL) is first proposed in \cite{John2020}, 3) Phase Q learning (PhaseQL) is studied in \cite{Kearns1998}.  But these algorithms are not studied with UCB based exploration policy. Further, it is not studied for RMAB. Our works is first to study these algorithm with UCB policy and extend to index learning algorithm in RMAB. 

\subsection{Our contributions}
Our contributions are as follows. 
We first discuss Q-learning algorithm and its faster variants---speedy Q- learning (SQL), generalized speedy Q-learning (GSQL) and phase Q-learning (PhaseQL). We present exploration exploitation schemes---$\epsilon$ greedy and upper confidence bound (UCB) approaches along with the Q-learning variants for action selection. We present a numerical example to illustrate the speed of convergence of these algorithms and observe that PhaseQL with UCB has the fastest rate of convergence, while Q-learning has slow convergence. 

 We study SQL, GSQL and PhaseQL for index learning with UCB as exploration-exploitation policy. The index learning algorithm is two-timescale stochastic approximations algorithm, where index is updated on slower timescale and Q-estimate updated on faster timescale. We use constant stepsize scheme.
We observe from numerical examples that PhaseQL with UCB has faster convergence than other algorithms.


Paper is organized as follows. In section \ref{sec:preliminaries}, we discuss preliminaries on Q-learning, its variants and provide faster variants of PhaseQL. We study restless-bandit model and index learning algorithm with different Q-learning approach in \ref{sec:RMAB-index-learning}.  We demonstrate performance of proposed learning algorithms using numerical example in section \ref{sec:numerical examples} and concluding remarks in section \ref{sec:conclusion}.

\section{Preliminaries on Q-learning and its variants}
\label{sec:preliminaries}
We consider finite state and finite action Markov decision process. 
Suppose that MDP has state $s \in \mathcal{S},$  action of the agent $a \in \mathcal{A}$ and agent received reward $r = r(s,a).$ The state of an environment changes to $ s^{\prime}$ from state $s$ under action $a$ according to $p_{i,j}^a.$ The transition probability matrix is given by $P^a = [[p_{i,j}^a]].$ The agent interact with an environment through sequence of actions. The goal of agent  is to maximize the long term cumulative discounted reward function

\begin{equation}
   V^{*}(s) = \max_{\pi} \mathbb{E}\left[ \sum_{t=0}^{\infty} \beta^t r_t \vert s_0 =s,  \pi  \right].
\end{equation}

 The optimal policy is stationary and an optimal dynamic program is given by 
\begin{eqnarray*}
    Q^*(s,a) &=& r(s,a) + \beta \sum_{s^{\prime} \in \mathcal{S} } p_{s,s^{\prime}}^a  
   \max_{a^{\prime}} Q^*(s^{\prime},a^{\prime}), 
 \\
V^{*}(s) &=& \max_{a \in \mathcal{A}} Q^*(s,a).
\end{eqnarray*} 

We assume that reward matrix $R =[[r(s,a)]]$ and probability matrix $P^a,$ $a \in \mathcal{A}$ are unknown. In \cite{Watkins1989}, Q-learning algorithm is proposed which an incremental learning scheme. Here,  the agent experience the  data: in the $n^{th}$ time step, the agent  observes its current state $ s,$ selects  an action $ a,$ 
observes the next state $ s^{\prime},$ and receives an immediate reward $r_n = r(s,a).$ Then agent updates its estimate on Q-function (state-action value) at time step $n+1$ using previous Q-value estimate $Q_{n}$ according to: 
\begin{eqnarray}
Q_{n+1}(s, a) = Q_{n}(s, a) + \alpha_n \left[r_n + \beta  \max_{a^{\prime}} Q_{n}(s^{\prime},a^{\prime}) 
\right. \nonumber \\ \left.
- Q_{n}(s, a)  \right] 
\end{eqnarray}
for  
$(s,a) \in \mathcal{S} \times \mathcal{A},$ otherwise
$Q_{n+1}(s,a) = Q_n(s,a).$
Here,  $\alpha_n$ is learning rate. 
This is also referred as asynchronous Q-learning, only one state-action pair is updated at a given time and updates are asynchronous. 
From \cite{Watkins1989,Jaakkola1993,Borkar08}, Q-learning algorithm converges to desired Q-value function if  for each state-action $(s,a) \in \mathcal{S} \times \mathcal{A},$ the $Q_n(s,a)$ estimate are updated infinitely often as $n \rightarrow \infty.$  




We next describe variants of Q-learning algorithms.  

\subsection{Speedy Q Learning (SQL)} 
We discus speedy Q-learning algorithm from \cite{Azar2011}. It is studied to resolve slow convergence of classical Q-learning algorithm which uses two-successive estimate of state-action value function.  For given state $s,$ the agent takes an action $a$ and reward $ r(s,a)$ and next state $s^{\prime}$ is observed by the agent. The agent maintains the two successive estimate for Q-function. $Q_{n-1}$ and $Q_{n}$ are estimate of Q-values at the timesteps $n-1$ and $n.$ 
Define the operator $\Gamma$ 
\begin{eqnarray}
    \Gamma Q(s, a) := r + \beta  \cdot \max_{a'} Q(s', a').
\end{eqnarray}
For each $(s,a) \in \mathcal{S} \times \mathcal{A},$ Q-learning update at timestep $n+1$ is as follows. 
\begin{eqnarray*}
  Q_{n+1}(s, a) := Q_n(s, a) + \alpha_{n} \cdot \left( \Gamma  Q_{n-1}(s, a) - Q_n(s, a)\right) + \nonumber \\ (1 - \alpha_n) \cdot \left(\Gamma Q_n(s, a) - \Gamma Q_{n-1}(s, a)\right), 
\end{eqnarray*}  
where $\alpha_n = \frac{1}{n+1}.$

\subsection{Generalized speedy Q Learning (GSQL)}  
The successive relaxation of speedy Q-learning is called as  generalized speedy Q-learning and it is proposed in \cite{John2020}. The relaxation parameter $w$ is introduced in Bellman operation.
We define operator $\widehat{T}$ as follows. 
\begin{eqnarray*} 
\widehat{T} Q(s, a) := w \cdot r + (1-w + \beta \cdot w) \cdot \max_{a'} Q(s', a').
\end{eqnarray*}
Here, there two successive estimate of Q-function is required. Then the Q-learning update at timestep $n+1$ is given by
\begin{eqnarray*}
Q_{n+1}(s, a) := Q_n(s,a)+ \alpha_n \cdot \left( \widehat{T} Q_{n-1}(s, a) - Q_n(s, a)\right)  \nonumber \\ +  (1- \alpha_n)\cdot \left(  \widehat{T} Q_n(s, a) -  \widehat{T} Q_{n-1}(s, a)\right).
\end{eqnarray*} 
The parameter $w$ can be optimized and it is function of transition probability matrix. 
it depend on probability of staying in same state. Further $w \geq 1.$ 

\subsection{Phase Q Learning (PhaseQL)}
In this algorithm,  for given state-action pair $(s,a),$  $m$ samples of next state are drawn randomly and it is used to provide estimate of transition probabilities and combine previous Q-estimate. If $m$ is large, then $\frac{1}{m}\sum_{i=1}^{m} 1_{(s,a,s_i^n= s^{\prime})} \approx p_{s,s^{\prime}}^a.$ Inspired from these the phase Q-learning algorithm is given as follows.
\begin{eqnarray*}
    Q_{n+1}(s,a) =  r(s,a) + \beta \frac{1}{m} \sum_{i=1}^{m}  \left[   V_{n}(s_i^{n})  \right], 
\end{eqnarray*}
and 
\begin{eqnarray*}
    V_{n+1}(s) = \max_{a}  Q_{n+1}(s,a).
\end{eqnarray*} 
Here, $\{ s_k^{n} \}_{k=1}^{m}$  are the states observed from $(s,a)$ in $m$ samples during $n$th time step.  
This is inspired from monte-carlo sampling approach. It is first introduced in \cite{Kearns1998}. The phase Q-learning algorithm is not an incremental scheme, as classical $Q$-learning is incremental scheme. Advantage of this algorithm is faster learning rate that classical Q-learning.  It is also called as empirical Q-learning and it is analyzed in \cite{Kalathil2021}. The phase $Q$ learning algorithm is modified and used in study of variance reduced value iteration algorithm, \cite{Sidford2023}.   



The intuition for faster convergence of algorithm: In each phase, there are $m$ samples of future state available from given state and action. It is used in next estimate to form Q value from preceding estimate of Q function. This provides advantages  as it is explores other state values frequently.

\subsection{Action selection policy: $\epsilon$-greedy and UCB policy} 
We discuss the action selection policies--- $\epsilon-$greedy policy and UCB policy. The convergence speed of algorithm futher depends on exploration-exploitation scheme used during Q-learning. 
In the $\epsilon-$greedy policy, the action at each step  is selected according to greedy policy with probability $1-\epsilon$ which uses $Q$ learning estimate  and and action is selected randomly with probability $\epsilon.$ 

\begin{eqnarray*}
    a_n =  
    \begin{cases}
        a & \mbox{with prob. $\frac{\epsilon}{|\mathcal{A}|}$}, \\
        \arg\max_{a \in \mathcal{A}} Q_{n+1}(s,a) & \mbox{with prob. $1-\epsilon.$}
    \end{cases}
\end{eqnarray*}

In the UCB based action selection scheme, action is selected according to the following rule.
\begin{eqnarray*}
    a_n = \arg\max_{a} \left( Q_n(s, a) + c \cdot \sqrt{\frac{\log(n+1)}{N(s, a) +1}} \right).
\end{eqnarray*}
Observe next state $s^{\prime}$ and reward $r.$ Then Q-learning algorithm is estimate as follows.
\begin{eqnarray*}
    Q_{n+1}(s, a) &=& Q_{n}(s, a) + \alpha \cdot \left( r + \beta  V_n(s^{'}) - Q_n(s, a) \right) \\
    V_n(s) &=& \min \left( V_{max}, \max_{a^{'}} Q_n(s, a^{'}) \right).
\end{eqnarray*} 
Define $s \leftarrow s'$ and repeat the algorithm.

This UCB based exploration-exploitation scheme can be utilised in SQL, GSQL and Phase QL for action selection. 

\subsection{Numerical example with no structure on transition model}
\label{sec:numerical-example-sab}
This example has five states and there is no structural assumption in the transition probability matrices. However, there is one for the reward matrix. For passive action ($a = 0$), the reward is linearly increasing along the states, and for active action ($a = 1$), the reward is linearly decreasing along the states.
\begin{eqnarray*}
  P_0 &=&  \begin{bmatrix}
    0.1502 & 0.0400 & 0.4156 & 0.0300 & 0.3642 \\
    0.4000 & 0.3500& 0.0800& 0.1200& 0.0500 \\
    0.5276 & 0.0400 &0.3991 & 0.0200 & 0.0133  \\
    0.0500 & 0.1000& 0.1500& 0.2000& 0.5000 \\
    0.0191 & 0.0100 & 0.0897 & 0.0300 & 0.8512 
\end{bmatrix}, \\
 P_1 &=& 
\begin{bmatrix}
    0.7196 & 0.0500 & 0.0903 & 0.0100 & 0.1301 \\
    0.5500 & 0.2000& 0.0500& 0.0800& 0.1200 \\ 
    0.1903 & 0.0100 & 0.1663 & 0.0100 & 0.6234 \\
    0.2000 & 0.0500 &0.1500 & 0.1000 & 0.5000 \\
    0.2501 & 0.0100& 0.3901 & 0.0300 & 0.3198 
\end{bmatrix}, \\
R &= &
\begin{bmatrix}
    0.4580 & 0.9631  \\
    0.5100 & 0.8100 \\
    0.6508 & 0.7963  \\
    0.6710 & 0.6061 \\
    0.6873 & 0.5057  
\end{bmatrix}.
\end{eqnarray*}
In this example we used the discount factor $\beta = 0.9$, step-size $\alpha = 0.02$, number of iterations $T_{\max}= 30000$ and exploration factor  $\epsilon = 0.3.$ In Phase QL, we use $m=20.$

\begin{figure}
  \begin{center}
    \begin{tabular}{cc}
      \includegraphics[scale=0.23]{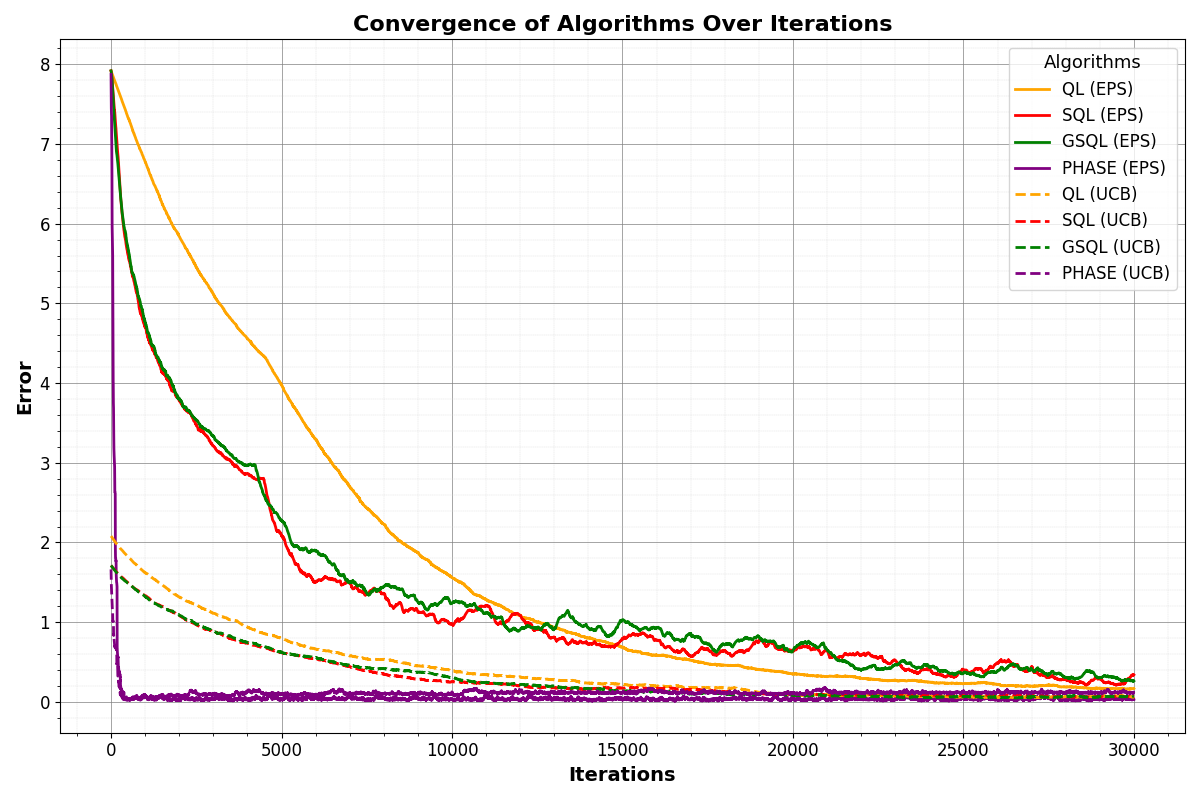}
    \end{tabular}
  \end{center}
  \caption{Performance of different Q-learning algorithms}
  \label{fig:Q-learning-variants-no-structure-model}
\end{figure} 
We plot error  $e_n = \mathrm{mean}(|Q_n(s,a,) - Q^*(s,a)|)$ as function of iteration $n.$
From Fig.~\ref{fig:Q-learning-variants-no-structure-model}, we observe that QL-UCB, SQL-UCB, GSQL-UCB and PhaseQL-UCB have higher rate of convergence than that with $\epsilon$ greedy exploration policy.  PhaseQL outperform QL,SQL, GSQL and having fastest convergence rate.  

\begin{figure}
  \begin{center}
    \begin{tabular}{cc}
      \includegraphics[scale=0.32]{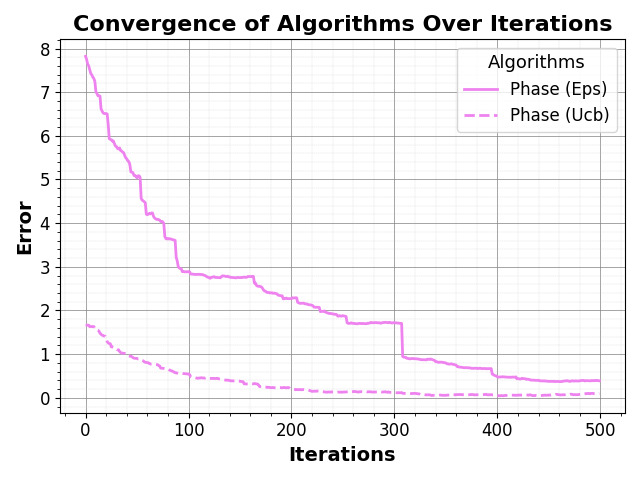}
    \end{tabular}
  \end{center}
  \caption{Performance of Phase Q-learning with $\epsilon$ greedy abd UCB algorithms}
  \label{fig:PhaseQ-learning-variants-no-structure-model}
\end{figure}

In Fig.~\ref{fig:PhaseQ-learning-variants-no-structure-model}, we plot Phase Q Learning with $\epsilon$-greedy and UCB algorithms. Note that PhaseQL with UCB has faster convergence than that of $\epsilon$-greedy. We further note that this can speed up the index learning for RMAB.  


\section{Restless Bandits Formulation}
\label{sec:RMAB-index-learning}
Consider restless multi-armed bandit with finite  arms $N.$ and finite states $K.$ Each arm is modeled as independent Markov decision processes with finite states $K$ and two actions. Let $\mathcal{S}$ be the state space for each arm, $X_t^i$ denotes the state at the beginning of time slot $t$ for arm $i$ and $X_t^i \in \mathcal{S}.$ The action of arm $i$ at time $t$ is denoted by $a_t^i$ with following interpretation 
\begin{eqnarray}
    a_t^i = 
    \begin{cases} 
    1 & \mbox{If arm $i$ is played, } \\ 
    0 & \mbox{If arm $i$ not  played.}
    \end{cases} 
\end{eqnarray}
The state dynamics of arm $i$ changes according to transition probabilities that depends on action $a^i.$ Let $[[p_{s,s^{\prime}}^{a^i}]]$  be the transition probability matrix for arm $i$  and action $a^i \in \{1,0\}.$ Reward from state $s$ under action $a^i$ for arm $i$ is $r_i(s,a^i).$ The discounted cumulative expected reward is 
\begin{eqnarray}
\max_{\pi} \mathbb{E}^{\pi}\left[ \sum_{t=1}^{\infty} \beta^{t-1} \left(\sum_{n=1}^{N} r_i(X_t^i,a_t^i) \right) \right] 
\end{eqnarray}
subject to constraint: $\sum_{n=1}^{N} a_t^i = M.$  At each instant $M$ arms are played. 

After  Whittle relaxation of the constraints, we get the new constraint: 
\begin{eqnarray}
    \sum_{t=1}^{\infty} \beta^{t-1} \left(\sum_{n=1}^{N} a_t^i \right) = \frac{M}{(1-\beta)}.
\end{eqnarray}
Next Lagrangian relaxation of optimization problem is as follows. 
\begin{eqnarray}
  \max_{\pi}  \mathbb{E}^{\pi}\left[ \sum_{t=1}^{\infty} \beta^{t-1} \left( r_i(X_t^i,a_t^i)  +  \lambda (1- a_t^i) \right) \right] .
\end{eqnarray}
We can rewrite the optimization problem: 
\begin{eqnarray}
   \max_{\pi} \mathbb{E}^{\pi}\left[ \sum_{n=1}^{N} \left( \sum_{t=1}^{\infty} \beta^{t-1}  \left( r_i(X_t^i,a_t^i)  +  \lambda (1- a_t^i \right) \right)  \right] . 
\end{eqnarray}
We can solve the problem for each arm separately. This is equivalent to solving $N$ independent armed single-armed bandit problem. Thus single-armed bandit problem is 
\begin{eqnarray}
  \max_{\pi}  \mathbb{E}^{\pi}\left[ \sum_{t=1}^{\infty} \beta^{t-1}  \left( r_i(X_t^i,a_t^i)  +  \lambda (1- a_t^i \right)  \right] .
\end{eqnarray}

\subsection{Single-armed restless bandit problem}
For notation simplicity we remove superscript $i.$ We consider single armed bandit and since rewards are bounded and then the optimal dynamic programming equation is as follows. 
\begin{eqnarray}
    V(s) = \max_{a \in \{0,1\}} \{ r(s,a) + \lambda (1-a) + \beta \sum_{s^{\prime}} p_{s,s^{\prime}}^{a} V(s^{\prime})\}.
\end{eqnarray}
The state-action value, i.e., Q-functions is given as follows. 
\begin{eqnarray*}
    Q^{\lambda}(s,a) =  r(s,a) + (1-a)\lambda +   \beta \sum_{s^{\prime} \in \mathcal{S} } p_{s,s^{\prime}}^a  
   \max_{a^{\prime}} Q^{\lambda}(s^{\prime},a^{\prime}),
\end{eqnarray*}
\begin{eqnarray*}
    V(s) &=& \max_{a \in \mathcal{A}} Q(s,a).
\end{eqnarray*}
Here, $\lambda$ is Lagrangian multiplier which is also called as subsidy given for not playing action. 

The Whittle index is the minimum subsidy requires to hold the state-action value function to be equal for both actions, that is, there is $\lambda$ such that $Q^{\lambda}(s,a=1) - Q^{\lambda}(s,a=0) =0.$ 
The index $W(s) = \min \{ \lambda:  Q^{\lambda}(s,a=1) - Q^{\lambda}(s,a=0) =0 \}$ for $s \in \mathcal{S}.$ This is applicable when model is known.
Our interest here is to study Q-learning  based algorithm for fixed subsidy $\lambda$  and it is given as follows.  
\begin{eqnarray}
Q_{n+1}^{\lambda}(s, a) = Q_{n}^{\lambda}(s, a) + \alpha_n \left[r(s,a) + \lambda (1-a) + 
\right. \nonumber \\ \left. 
\beta  \max_{a^{\prime}} Q_{n}^{\lambda}(s^{\prime},a^{\prime}) 
- Q_{n}^{\lambda}(s, a)  \right] 
\end{eqnarray}
for  
$(s,a) \in \mathcal{S} \times \mathcal{A},$ otherwise
$Q_{n+1}^{\lambda}(s,a) = Q_n^{\lambda}(s,a).$  Here $\alpha_n$ is learning rate (stepsizes). Suppose the index (subsidy) is known to decision maker but the model is unknown, then running Q-learning algorithm one can learn the optimal Q-state value function. This is equivalent to solving Q-learning algorithm for MDP. 
This algorithm can be extended to other variant of Q-RL algorithms (SQL, GSQL and Phase-QL) with $r= r(s,a) + \lambda (1-a).$

\subsection{Two-timescale index learning algorithm}
In this section, we assume that index $\lambda$ is not known to the decision maker.
The index is function of state $\tilde{s},$  thus one requires to learn index $\lambda$ for each $\tilde{s} \in \mathcal{S}.$ We define
$Q^{*,\lambda}(s,a,\tilde{s})$ optimal Q-function for $(s,a,\tilde{s})$ and it is dependent on $\lambda.$ For fixed $\lambda$ and threshold state $\tilde{s},$ we have $Q^{*,\lambda}(\tilde{s},a=1,\tilde{s}) = Q^{*,\lambda}(\tilde{s},a=1,\tilde{s}).$ 
Let
$Q_n^{\lambda}(s, a, \tilde{s})$ be  learning estimate of Q function at time $n$ for  $(s,a,\tilde{s}).$ We expect that for large $n$ we have $Q_n^{\lambda}(s, a, \tilde{s}) \rightarrow  Q^{*,\lambda}(s,a,\tilde{s}).$ Hence for large $n,$ and at $\tilde{s}$ is a threshold state such that  $|Q_n^{\lambda(\tilde{s})}(\tilde{s},a = 1, \tilde{s}) - Q_n^{\lambda(\tilde{s})}(\tilde{s},a = 0, \tilde{s})|
\leq \delta$ for $0<\delta <<1.$ 

\begin{algorithm}
    \KwIn{Step-sizes, $\alpha, \gamma \in {(0, 1]}$; small $\epsilon > 0$; discount factor, $\beta$; threshold for convergence, $\delta$;  $T_{\max}$ and  $K_{\max}$}
    \KwOut{Indices, $\lambda(\tilde{s}),$ $\tilde{s} \in \mathcal{S}$}

    \textbf{Initialization}: $Q(s, a, \tilde{s})$ for all $s, \tilde{s} \in \mathcal{S}, a \in \{0,1\}$; $\lambda(s)$ for all $s \in \mathcal{S}$ 

    \For {$k = 0, 1, 2, \cdots, K_{\max}-1 $}
    {
        \For {$\tilde{s} = 0, 1, \cdots, S - 1$}
        {
            $q(\cdot , \cdot) \leftarrow Q(\cdot , \cdot , \tilde{s})$
            
            \For {$n = 0, 1, 2, \cdots, T_{\max}-1 $}
            {   
                 \textbf{Action:} $a= \mathrm{EEpolicy}(s,a, Q)$ \\
                \textbf{Observations:} $(r,s^{\prime}) = \mathrm{Observe}(s,a, \lambda(\tilde{s}))$ \\ 
                $r = r(s,a) + (1-a) \lambda(\tilde{s})$ \\
                \textbf{Q-RL update rule:} $q(s,a) = \mathrm{QRLupdate}(s,a,r,s^{\prime},\beta, \alpha)$ \\ 
                $s \leftarrow s'$
            }

            $Q^{\lambda(\tilde{s})}(\cdot,\cdot,\tilde{s}) \leftarrow q(\cdot,\cdot)$ 
        }

        \vspace{5pt}
        \For {$\tilde{s} = 0, 1, \cdots, S - 1$}
        {
            $\lambda(\tilde{s}) \leftarrow \lambda(\tilde{s}) + \gamma  \times  (Q^{\lambda(\tilde{s})}(\tilde{s}, 1, \tilde{s}) - Q^{\lambda(\tilde{s})}(\tilde{s}, 0, \tilde{s}))$
        }

        \vspace{5pt}
        \If {$\max_{\tilde{s}}|Q^{\lambda(\tilde{s})}(\tilde{s}, 1, \tilde{s}) - Q^{\lambda(\tilde{s})}(\tilde{s}, 0, \tilde{s})| < \delta$}{
            $\mathrm{break}$
        }
    }

    \Return{Indices, $\lambda(\tilde{s}),$ $\tilde{s} \in \mathcal{S}$} 
    \caption{Index Learning Algorithm}
    \label{algo:Index-learning}
\end{algorithm}
Learning of index $\lambda(\tilde{s})$ involves two loops in our Algorithm~\ref{algo:Index-learning}. 
In the inner loop, assuming subsidy $\lambda(\tilde{s})$ is fixed for each threshold state $\tilde{s} \in \mathcal{S},$  Q-learning update is performed  and stored in $Q(\cdot, \cdot, \tilde{s}).$  In the outer loop index $\lambda_n(\tilde{s})$ is updated as follows.
\begin{eqnarray*}
    \lambda_{n+1}(\tilde{s}) = \lambda_n(\tilde{s}) + \gamma (Q(\tilde{s}, 1, \tilde{s}) - Q(\tilde{s}, 0, \tilde{s})). 
\end{eqnarray*}
Here, $\gamma$ is constant step size.

This algorithm can be viewed as two timescale algorithm in which subsidy $\lambda_n(\tilde{s})$ is updated on slower timescale and the Q-learning update (Q-RL, SQL, GSQL, Phase-QL) is performed on faster timescale.  Here we consider constant stepsizes in both Q-learning update and index-learning. for given threshold state $\tilde{s} \in \mathcal{S},$ we have  
\begin{eqnarray}
    Q_{n+1}^{\lambda}(s, a, \tilde{s}) =  Q_{n}^{\lambda}(s, a, \tilde{s})  + \alpha \left[r(s,a) + \lambda_n(\tilde{s}) (1-a)+ \nonumber 
    \right. \\ \left. 
    \beta \max_{a^{'}} Q_{n}^{\lambda}(s, a^{'}, \tilde{s}) - Q_{n}^{\lambda}(s, a, \tilde{s})
    \right]
    \label{eqn:Q-RLupdate}
\end{eqnarray} 
\begin{eqnarray}
    \lambda_{n+1}(\tilde{s}) = \lambda_n(\tilde{s}) + \gamma  \left[Q_n^{\lambda}(\tilde{s}, 1, \tilde{s}) - Q_n^{\lambda}(\tilde{s}, 0, \tilde{s})\right]. 
    \label{eqn:index-update}
\end{eqnarray}
The iterative  algorithms are governed by two step-sizes ($\alpha$ and $\gamma$), where $\alpha$ and $\gamma$ are used in Q-learning and index learning, respectively. We terminate the algorithm in either of the following conditions: either the number of iterations exceeds the pre-defined maximum number of iterations, or the maximum difference, i.e., $\max_{\tilde{s}}|Q_n^{\lambda}(\tilde{s}, 1, \tilde{s}) - Q_n^{\lambda}(\tilde{s}, 0, \tilde{s})|$ becomes smaller than a pre-defined threshold. 

The analysis of index learning with two-timescale constant stepsizes is skipped due to space constraint.
We next discuss different index learning using variant of Q-RL.
\subsubsection{Index learning with Q-RL} 
In this scheme Q-RL-Update is according to Eqn.~\eqref{eqn:Q-RLupdate}. Once it converges after sufficiently large iteration, index is computed from Eqn.~\eqref{eqn:index-update}.  This is described in Algorithm~\ref{algo:Index-learning}. 

\subsubsection{Index learning with SQL} 
In this scheme Q-RL-Update is according to following Eqn.
\begin{eqnarray}
Q_{n+1}^{\lambda}(s, a,\tilde{s}) := Q_n^{\lambda}(s, a,\tilde{s}) + \alpha TE_1 + (1 - \alpha) TE_2      
\end{eqnarray}
where 
\begin{eqnarray*}
    TE_1  &=&  \Gamma  Q_{n-1}^{\lambda}(s, a,\tilde{s}) - Q_n^{\lambda}(s, a,\tilde{s}), \\
    TE_2 &=&  \Gamma Q_n^{\lambda}(s, a,\tilde{s}) - \Gamma Q_{n-1}^{\lambda}(s, a,\tilde{s}), \\
    \Gamma Q^{\lambda}(s, a) &=& r + \beta   \max_{a'} Q^{\lambda}(s', a').
\end{eqnarray*}
Note that we have considered constant step size SQL learning.
Once it converges after sufficiently large iteration $T_{\max}$, index is computed from Eqn.~\eqref{eqn:index-update}.   
SQL is faster than Q-learning by tracking the action-value functions of two time steps $n$ and $n-1.$ It  incorporates two learning rates—one conservative and the other more aggressive. Further, SQL ensures that the Q-values converge almost surely to $Q^*$ with the rate $\sqrt{\frac{1}{T_{\max}}} $, where $T_{\max}$ is the number of time steps in SQL.

\subsubsection{Index learning with GSQL}

In this scheme Q-RL-Update is according to following Eqn.
\begin{eqnarray*}
Q_{n+1}^{\lambda}(s, a,\tilde{s}) := Q_n^{\lambda}(s,a,\tilde{s})+ \alpha TE_3   +  (1- \alpha) TE_4,
\end{eqnarray*} 
where
\begin{eqnarray*}
    TE_3 &=&   \widehat{T} Q_{n-1}^{\lambda}(s, a,\tilde{s}) - Q_n^{\lambda}(s, a,\tilde{s})  \\ 
    TE_4 &=&  \widehat{T} Q_n^{\lambda}(s, a,\tilde{s}) -  \widehat{T} Q_{n-1}^{\lambda}(s, a,\tilde{s}) \\
    \widehat{T} Q^{\lambda}(s, a,\tilde{s}) &=& w  r + (1-w + \beta  w)  \max_{a'} Q^{\lambda}(s', a',\tilde{s}).
\end{eqnarray*}

Once it converges after sufficiently large iteration $T_{\max},$ index is computed from Eqn.~\eqref{eqn:index-update}.    
GSQL uses the generalized Bellman operator. It is modification of SQL and relaxation with parameter $w \geq 1.$ The contraction factor of the generalized Bellman operator is lower than that of the standard Bellman operator $w=1.$ This improves  the finite-time performance bound and improves convergence rate. 


\subsubsection{Index learning with Phase-Q RL}

In this scheme Q-RL-Update is according to following Eqn.
\begin{eqnarray*}
    Q_{n+1}^{\lambda}(s,a,\tilde{s}) =  r(s,a) + \beta \frac{1}{m} \sum_{i=1}^{m}  \left[   V_{n}^{\lambda}(s_i^{n}),\tilde{s}  \right] 
\end{eqnarray*}
where
\begin{eqnarray*}
    V_{n}^{\lambda}(s_i^{n},\tilde{s}) = \max_{a^{\prime}}  Q_{n}^{\lambda}(s_i^{n},a^{\prime},\tilde{s}).
\end{eqnarray*} 
and $\{ s_i^{n} \}_{i=1}^{m}$  are states observed from $(s,a)$ in $m$ samples.

Once it converges after sufficiently large iteration $T_{\max},$ index is computed from Eqn.~\eqref{eqn:index-update}.
%





\section{Numerical Example}
\label{sec:numerical examples}

\begin{figure}
  \begin{center}
    \begin{tabular}{cc}
      \includegraphics[scale=0.45]{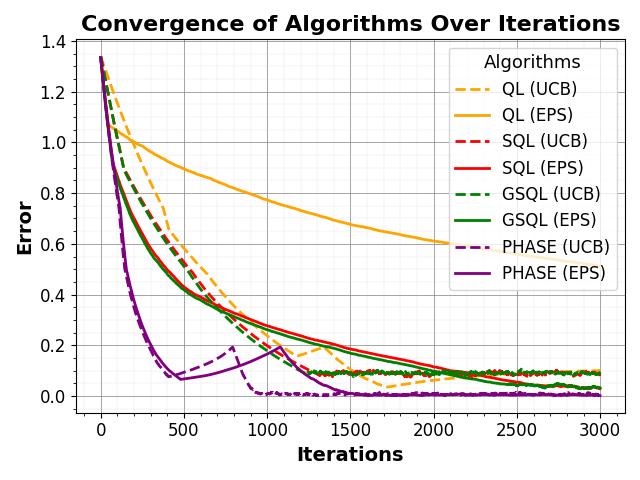}
    \end{tabular}
  \end{center}
  \caption{Index learning performance with different Q-learning algorithms}
  \label{fig:index-learning-variants-no-structure-model}
\end{figure}

We now demonstrate the performance of index learning algorithms with variants of Q-learning. We consider same example as discussed in section~\ref{sec:numerical-example-sab}. We use following parameter.  $\epsilon =0.3,$ $T_{\max} =10000,$ $K_{\max} = 3000,$  $m =20.$  $\alpha = 0.02$ and $\gamma= 0.005.$
We plot the error $E_n = \frac{1}{|\mathcal{S}|}\sum_{\tilde{s}  \in \mathcal{S}} |Q_n^{ \lambda}(\tilde{s}, 1, \tilde{s} ) - Q_n^{\lambda}(\tilde{s}, 0, \tilde{s})|.$
%
We plot $E_n$ as function of  iteration $n$. In Fig.~\ref{fig:index-learning-variants-no-structure-model}, we present the performance of index learning with Q-learning algorithms. We observe that PhaseQL (UCB and $\epsilon$ greedy) performs better than other Q-learning variants and have faster convergence rate. 
%

\section{Concluding Remarks}
\label{sec:conclusion}
We discussed the faster Q-learning algorithms.
We studied Whittle index learning algorithm with  Q-learning and its faster Q-learning variants---SQL, GSQL, PhaseQL. We also considered UCB and $\epsilon$-greedy as exploration policy for action selection. We observed from numerical example that PhaseQL with UCB has faster convergence compared to  Q-learning with $\epsilon$-greedy and UCB schemes.

\section{Acknowledgment}

The work of Rahul Meshram at IIT Madrad is supported by NFI Grant IIT Madras and SERB Project “Weakly Coupled POMDPs and Reinforcement Learning”. 
\bibliographystyle{IEEEbib}
\bibliography{its-references}

\end{document}